\pdfoutput=1

\documentclass[11pt]{article}

\usepackage[]{acl}

\usepackage{times}
\usepackage{latexsym}
\usepackage{color, colortbl}
\usepackage[T1]{fontenc}

\usepackage[utf8]{inputenc}

\usepackage{microtype}

\usepackage[utf8]{inputenc}
\usepackage{color}
\usepackage{tabularx}
\usepackage{tabu}
\usepackage{booktabs}
\usepackage{multirow}

\usepackage{graphicx} 
\usepackage{float}
\usepackage{subfigure} 
\usepackage{amssymb}

\usepackage{hyperref}
\usepackage{tablefootnote}
\usepackage{xspace}

\usepackage{float}
\usepackage{amsmath}
\usepackage{bm}
\usepackage{algorithmicx}
\usepackage{algorithm}
\usepackage{algpseudocode}

\usepackage{enumitem}

\definecolor{amber}{rgb}{1.0, 0.75, 0.0}
\newcommand{\draftonly}[1]{#1} 
\newcommand{\draftcomment}[3]{\draftonly{{\textcolor{#3}{[\textbf{#1--\textsc{#2}}]}}}}
\iffalse
    \newcommand{\yilun}[1]{\textcolor{blue}{}}
    \newcommand{\rui}[1]{}
\else
    \newcommand{\yilun}[1]{\textcolor{blue}{\bf\small [Yilun: #1]}}
    \newcommand{\rui}[1]{\draftcomment{#1}{rui}{amber}}
    
\fi

\newcommand{\ours}{\textsc{LoFT}\xspace}

\newcommand{\tabfact}{\textsc{TabFact}\xspace}

\newcommand{\logicnlg}{\textsc{LogicNLG}\xspace}

\title{\ours: Enhancing Faithfulness and Diversity for Table-to-Text \\ Generation via Logic Form Control}

\author{Yilun Zhao\thanks{~~Equal Contributions.}~~$^{1}$ \quad Zhenting Qi$^{*2}$ \quad Linyong Nan$^1$ \\ \bf{Lorenzo Jaime Yu Flores$^1$ \quad Dragomir Radev$^1$} \\
$^1$Yale University \quad $^2$ Zhejiang University \\
\texttt{yilun.zhao@yale.edu \quad zhenting.19@intl.zju.edu.cn}
}

\begin{document}
\maketitle
\begin{abstract}
Logical Table-to-Text (LT2T) generation is tasked with generating logically faithful sentences from tables. There currently exists two challenges in the field: 1) \emph{Faithfulness}: how to generate sentences that are factually correct given the table content; 2) \emph{Diversity}: how to generate multiple sentences that offer different perspectives on the table. This work proposes \ours, which utilizes logic forms as fact verifiers and content planners to control LT2T generation. Experimental results on the \logicnlg dataset demonstrate that \ours is the first model that addresses unfaithfulness and lack of diversity issues simultaneously. Our code is publicly available at \url{https://github.com/Yale-LILY/LoFT}.

\end{abstract}

\section{Introduction}
\begin{figure}[!t]
    \centering
    \includegraphics[width = 0.98\linewidth]{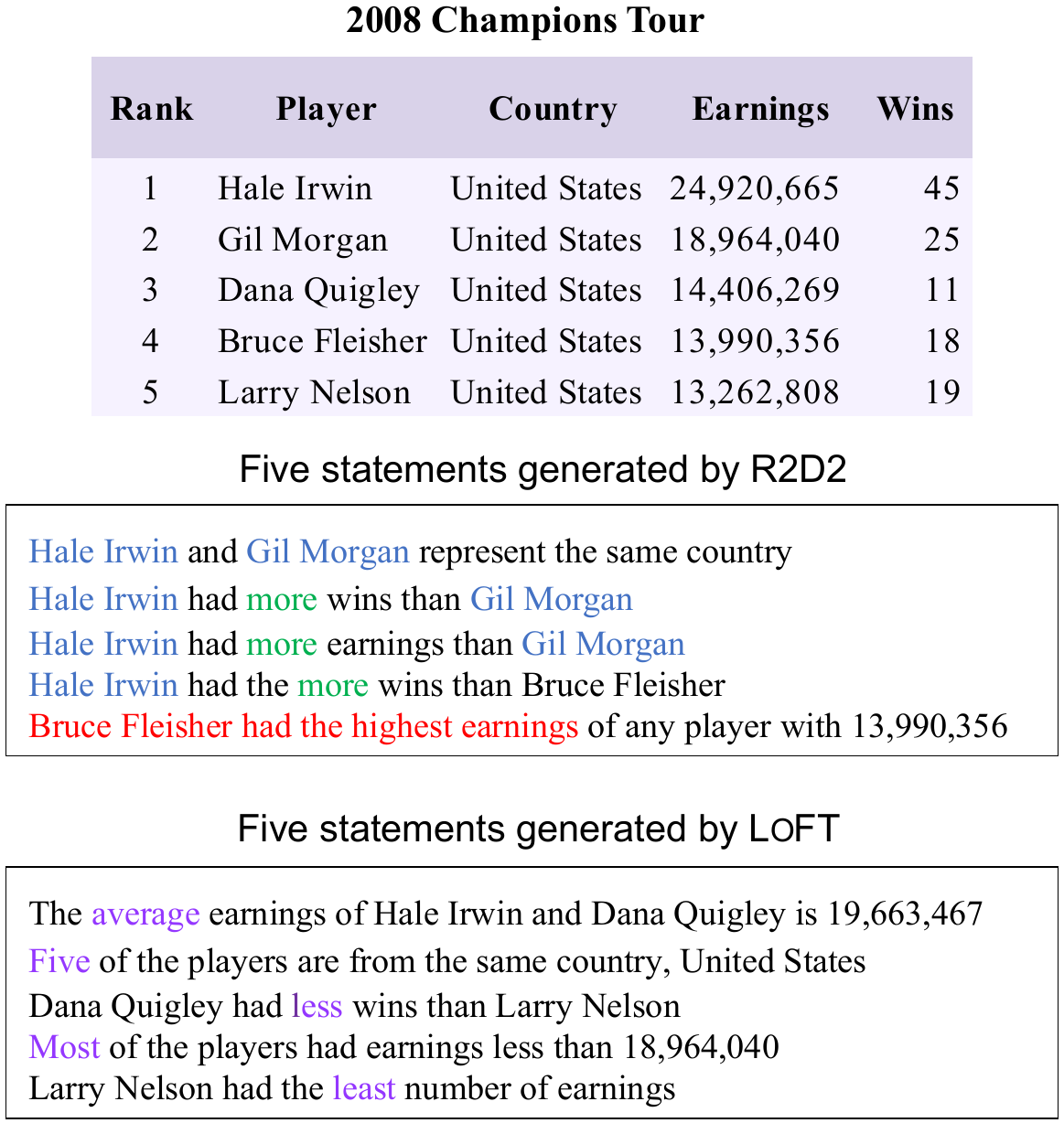}
    \caption{An example of logical table-to-text generation. (a) Statements generated by previous models~\cite{nan2022r2d2}: the generation suffers from 1) \emph{Lack of diversity}, as three of the generated statements are focused on the \textcolor[RGB]{51, 102, 255}{same table regions} (i.e., “Hale Irwin” and “Gil Morgan”), and three of them use the similar \textcolor[RGB]{0, 153, 51}{reasoning operations} (i.e., comparative); 2) \emph{Unfaithfulness}, as one of the generated statements is \textcolor{red}{factually incorrect} given the table content. (b) Statements generated by \ours: By utilizing logic forms to \emph{control} the generation, our method can generate multiple factually correct sentences that each use a \textcolor[RGB]{204, 153, 255}{different reasoning operation} to offer various perspectives on the tabular data.}
    \label{fig:example}
\end{figure}
Table-to-Text (T2T) generation aims to produce natural language descriptions from structured tables. 
A statement generated from tabular data can be inferred based on different levels of information (e.g., value of a specific cell, logical operation result across multiple cells).  Although current T2T models~\cite{lebret-etal-2016-neural, wiseman-etal-2017-challenges, puduppully-etal-2019-data, parikh-etal-2020-totto} have shown remarkable progress in fluency and coherence, they mainly focus on surface-level realizations without much logical inference.  

Recently, \citet{chen-etal-2020-logical} proposed \logicnlg, which is tasked with generating textual descriptions that require logical reasoning over tabular data (i.e., LT2T generation). 
LT2T generation is challenging as it requires a model to learn the logical inference knowledge from table-text pairs and generate multiple \emph{factually correct} sentences. 
Another challenge for LT2T generation is the \emph{diversity} of generated text. Natural Language Generation (NLG) encourages the diverse output of statements over a single input, as it provides various perspectives on the data and offers users more choices. In LT2T generation, requirements for diversity naturally emerge from the need to apply different logical operations to extract different levels of table information. 
However, current methods~\cite{chen-etal-2021-de, nan2022r2d2, liu2022plog, reastap} that address issues of unfaithfulness have overlooked the importance of diversity. As shown in Figure~\ref{fig:example}, multiple statements generated using current methods~\cite{nan2022r2d2} might only cover information from the same table region or logical operation. Such issues related to lack of diversity could limit the deployment of LT2T models in the real world. 

In this work, we attribute \emph{unfaithfulness} and lack of \emph{diversity} to the absence of \emph{controllability} over generation. 
Specifically, due to the large number of combinations of different logical operations and table regions,
the space of factually correct statements is exponentially large. However, \logicnlg uses the whole table as the input, without providing annotations related to any other explicit control attribute. As a result, it is hard and uncontrollable for neural models to decide a favorable choice of logical selections solely based on the table input. We believe such \emph{uncontrollability} leads to unfaithfulness and lack of diversity issues. 

This work proposes \ours, a framework that utilizes logic forms as mediators to enable \emph{controllable} LT2T generation. 
Logic forms~\cite{chen_2020_logic2text, Chen2020TabFact} are widely used to retrieve evidence and explain the reasons behind table fact verification~\cite{yang-etal-2020-program,yang-zhu-2021-exploring-decomposition, ou-liu-2022-learning}. In this work, logic forms are used as: 1) fact verifiers to ensure the factual correctness of each generated sentence; and 2) content planners to control which logical operation and table region to use during the generation. Experimental results show that \ours surpasses previous methods in faithfulness and diversity simultaneously.

\section{Related Work}
\paragraph{Logical Table-to-Text (LT2T) Generation} \logicnlg~\cite{chen-etal-2020-logical} is tasked with generating logically faithful sentences from tables. To improve the faithfulness of generated statements, 
\citet{nan2022r2d2} trained a system both as a generator and a faithfulness discriminator with additional replacement detection and unlikelihood learning tasks.
\citet{liu2022plog} pre-trained a model on a synthetic corpus of table-to-logic-form generation. 
\citet{reastap} demonstrated that faithfulness of LT2T can be improved by pre-training a generative language model over synthetic Table QA examples.
However, these methods overlook the importance of diversity in T2T generation, and might generate multiple statements that cover the same table regions or reasoning operations. 
Previous methods in NLG proposed to improve diversity by modifying the decoding techniques~\cite{li-etal-2016-diversity}. However, these approaches degrade faithfulness as measured against baselines~\cite{perlitz2022diversity}. 
To enable controllable generation and improve diversity, \citet{perlitz2022diversity} used logical types of statements as a control. However, such methods still suffer from problems related to unfaithfulness, and may generate statements covering limited table regions.
This work proposes to leverage the logic form as a fact checker and content planner to control LT2T generation, which tackles the challenges about faithfulness and diversity at the same time. 

\paragraph{Table Fact Verification via Logic Form}
Logic forms are widely used in Table Fact Verification~\cite{Chen2020TabFact}.  
Specifically, given an input statement, the model~\cite{yang-etal-2020-program,yang-zhu-2021-exploring-decomposition, ou-liu-2022-learning} will first translate it into logic form. Then the logic form will be executed over the table, and return \texttt{true/false} as the entailment label for a given statement.
While several works ~\cite{chen_2020_logic2text, shu-etal-2021-logic, liu2021improving} focused on generating fluent statements from logic forms, the utilization of logic forms to benefit LT2T generation is still unexplored.
\begin{figure*}[!t]
    \begin{minipage}[b]{0.48\linewidth}
            \centering
            \includegraphics[width = \linewidth]{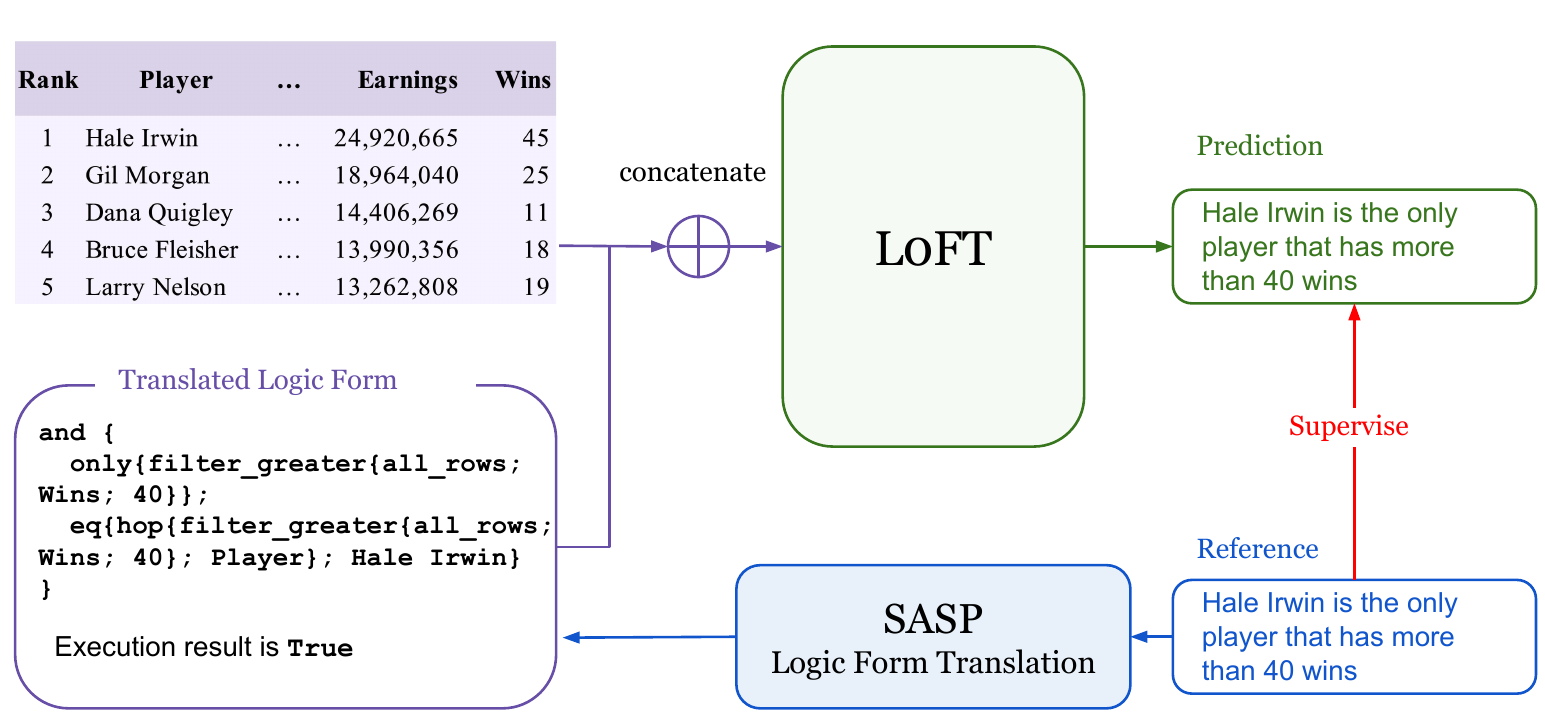}
            \caption*{(a) \ours training stage.}
    \end{minipage}\quad
    \begin{minipage}[b]{0.48\linewidth}
            \centering
            \includegraphics[width = \linewidth]{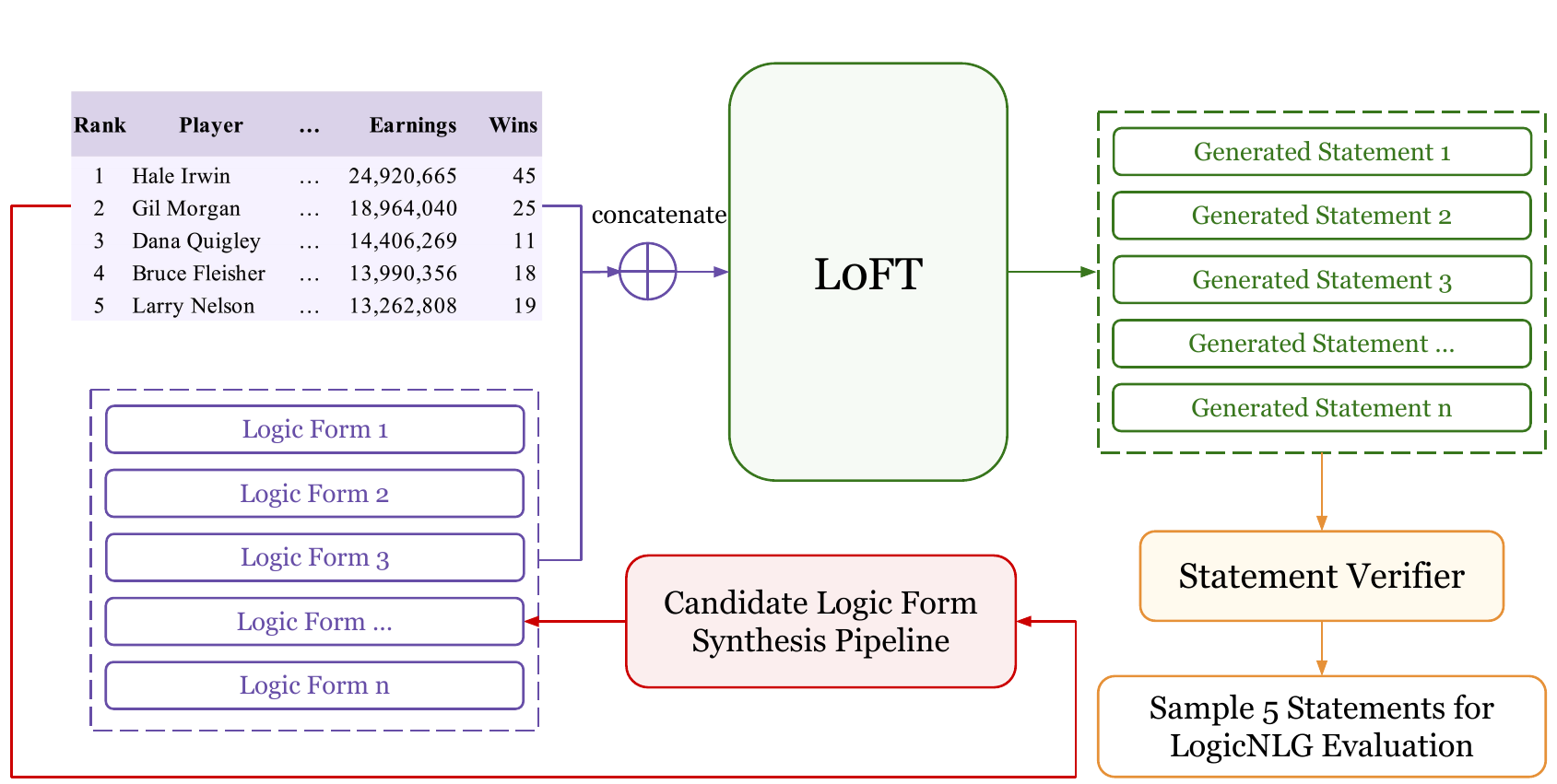}
            \caption*{(b) \ours inference stage.}
    \end{minipage}
    \centering
    \caption{The illustration of \ours. (a) During the training stage, the SASP model is first applied to translate each statement in the \logicnlg training set into the logic form. Then \ours is trained to generate the reference statement given the translated logic form and serialized table data. (b) During the inference stage, given each table, the logic form synthesis pipeline was first applied to synthesize candidate logic forms that cover different table regions and logical operations. \ours is applied to generate statements for each candidate logic form. Then a statement verifier is used to filter out those potentially unfaithful statements. As a result, \ours can generate a diverse set of faithful statements covering different table regions and reasoning operations. For each table in the \logicnlg test set, we randomly sampled five candidate statements for evaluation.}
    \label{example}
\end{figure*}

\section{\ours}
This section first introduces the logic form utilized, and then delves into the training and inference process of \ours. We also explain how the use of logic forms can enhance both faithfulness and text-diversity in LT2T generation.
\subsection{Logic Form Implementation}
Logic forms are widely used to retrieve evidence and explain the reasons behind table fact verification. We use the same implementation as \citet{chen_2020_logic2text}, which covers 8 types of the most common logical operations (e.g., count, aggregation) to describe a structured table. Each logical operation corresponds to several Python-based functions. For example, the definition of function \texttt{all\_greater(view, header, value)} under “majority” category is: checking whether all the values under \texttt{header} column are greater than \texttt{value}, with the scope (i.e., \texttt{view}) of all or a subset of table rows. The complete list of logical operation types and corresponding function definitions are shown in Table~\ref{table:app_func} in Appendix.

\subsection{\ours Training}
\paragraph{Training Task Formulation}
Given the serialized tabular data with selected columns as $T$, the training objective of \ours is to generate a sentence $\boldsymbol{y}=(y_1, y_2, \dots, y_n)$ that is both fluent and faithful, with the translated logic form $l$ as control.
\begin{equation}
   \boldsymbol{y} = \mathrm{argmax}\prod_{i=1}^n P(y_i | y_{<i}, T,\,l;\, \theta)
    \label{eq:obj}
\end{equation}
where $\theta$ denotes the parameters of a seq2seq LM. 

\paragraph{Training Dataset Collection} \label{sec: train_data}
Since the \logicnlg dataset does not contain logic form annotations, we had to augment each statement in the training set with its corresponding logic forms. To construct \textit{\{statement, logic form\}} parallel data for the \logicnlg training set, we adapted SASP~\cite{ou-liu-2022-learning}, the state-of-the-art model for \tabfact dataset, which leverages structure-aware semantic parsing over tables to translate the given statement into logic form. In this work, given an example in the \logicnlg training set, SASP was applied to generate its logic form, resulting in a total of 15,637 examples for \ours training.

\subsection{\ours Inference}
During the inference stage, for each given table, we first applied the logic form synthesis pipeline to synthesize multiple candidate logic forms~\cite{liu2022plog}. For each of these logic forms, the system generates its corresponding statement. The faithfulness of these statements were further checked by a verifier.

\paragraph{Logic Form Synthesis Pipeline} \label{sec: synthesis}
To synthesize a candidate set of logic forms paired with each supporting table, we applied a similar logic form synthesis pipeline as \citet{liu2022plog}. 

We extracted templates of logic forms from the collected \ours training dataset. Specifically, we categorized functions with similar definitions (e.g., \texttt{max/min}, \texttt{greater/less}) into smaller groups to obtain a more abstract template. Each function category corresponded to one unique table reasoning skill. For each template, we masked specific entities in the logic forms as typed placeholders (i.e., \texttt{col} to denote a column header, \texttt{obj} to denote an object). Finally, we obtained 45 different templates, covering 8 table logical operations. Table \ref{table:app_func} shows the complete list of reasoning 
operations and corresponding function definitions.

Given the table and each set of selected columns, the pipeline would synthesize a total of 20 candidate logic forms whose execution result over the table is \texttt{True}. 
To generate a candidate logic form, the pipeline first sampled a logic form using a weighted-sampling technique with the weight equal to the template distribution in the \ours training dataset (Section \ref{sec: train_data}). The weighted sampling is to ensure that the generated candidate logic forms follow a similar distribution as \logicnlg. To instantiate the sampled template, a bottom-up sampling strategy is adopted to fill in each placeholder of the template and finally generate the logic form.

\paragraph{Statement Generation \& Verification}\label{sec:verifier}
Through the logic form synthesis pipeline, we obtained a large number of candidate logic forms. For each logic form, we used \ours to generate the corresponding statement. The candidate statements might still contain some factually incorrectness, thus we applied an NLI-based verifier to filter out those potentially unfaithful generations. 
Specifically, we used the \textsc{TabFact}~\cite{Chen2020TabFact} dataset to train a classifier, which adopts RoBERTa-base as the backbone. We fed each generated statement and its corresponding table into the classifier, and only kept those statements that were predicted as entailed. Then we randomly sampled five statements as the output for each table in \logicnlg.

\subsection{Enhancing LT2T via Logic Form Control}
This subsection provides two perspectives to explain why logic forms can help improve both faithfulness and diversity of LT2T generation.
\paragraph{Logic Form as Content Planner}
Logic forms pass column or cell values as arguments, guiding the model to focus on relevant table regions. The function category of the logic form, such as \texttt{count}, helps the model better organize logical-level content planning.

\paragraph{Logic Form as Fact Verifier}
Logic forms are defined with unambiguous semantics, hence are reliable mediators to achieve faithful and controllable logical generations. During the inference stage, we synthesize candidate logic forms with 100\% execution correctness. The sampled logic form serves as a fact verifier and conveys accurate logical-level facts for controllable LT2T generation.

\begin{table*}[!t]
\centering
\resizebox{1.98\columnwidth}{!}{
\begin{tabular}{l|c|cc|ccc}
\toprule
\multirow{2}{*}{Model} & \multicolumn{1}{c|}{Surface-level} & \multicolumn{2}{c|}{Diversity-level} & \multicolumn{3}{c}{Faithfulness-level}\\
\cmidrule(lr){2-2}  \cmidrule(lr){3-4} \cmidrule(lr){5-7} 
& BLEU-1/2/3$\uparrow$ & Distinct-2$\uparrow$  & s-BLUE-4$\downarrow$ & SP-Acc$\uparrow$  & NLI-Acc$\uparrow$  & TAPEX-Acc$\uparrow$ \\
\midrule
GPT2-TabGen~\cite{chen-etal-2020-logical} & 48.8/27.1/12.6 & 59.0 & 55.3  & 42.1 & 68.7 & 45.0\\
GPT2-C2F~\cite{chen-etal-2020-logical} & 46.6/26.8/13.3 & 60.3 & 52.8 & 42.7 & 72.2 & 44.1\\
DCVED$^*$~\cite{chen-etal-2021-de} & 49.5/28.6/15.3 &  -- & -- & 43.9 & 76.9 & -- \\
DEVTC$^\ddag$~\cite{perlitz2022diversity} & 51.3/30.6/16.3 & 73.7 & 21.3  & 44.3 & 77.9 & 55.6 \\
R2D2~\cite{nan2022r2d2} & 51.8/32.4/18.6 & 60.1 & 51.5 &  50.8 & 85.6 & 60.2\\
\midrule
\ours & 48.1/27.7/14.9  & \textbf{79.5} & \textbf{17.7} & \textbf{57.7} & \textbf{86.9} & \textbf{61.8} \\

\bottomrule
\end{tabular}
}
\caption{Performance on the \logicnlg test set. $\ddag$: results from our own implementation; $*$: code not released and we used the results reported in original papers. \ours achieves great improvement on faithfulness and diversity.}
\label{tab:logicnlg-auto}
\end{table*}
\begin{table}[!t]
\resizebox{1\columnwidth}{!}{
\renewcommand{\tabcolsep}{0.5mm}
    \centering
    \small
    \begin{tabular}{l|cc|cc|cc}
    \toprule
        Diversity & \multicolumn{2}{c|}{DEVTC} & \multicolumn{2}{c|}{R2D2} & \multicolumn{2}{c}{\ours} \\
        Criteria & Best$\uparrow$ & Worst$\downarrow$ & Best$\uparrow$ & Worst$\downarrow$ & Best$\uparrow$ & Worst$\downarrow$ \\ \midrule
        Table Coverage &  8 & 16 & \cellcolor{red!20} 5 & \cellcolor{red!20}20 & \cellcolor{green!20} 29 &  \cellcolor{green!20}5 \\
        Reasoning Op& 19 & \cellcolor{green!20}1 & \cellcolor{red!20} 2 & \cellcolor{red!20}37 &  \cellcolor{green!20} 24  & 2 \\
    \bottomrule
    \end{tabular}
}
\caption{Number of times the system was selected as best or worst by majority vote (including ties). \ours outperforms other baselines in terms of diversity for both table coverage and reasoning operations.}
\label{tab:human-study-vanilla}
\end{table}
\begin{table}[!t]
\centering
\small
\begin{tabular}{l|c|c}
\toprule
\multirow{2}{*}{Model} & Faithfulness $\uparrow$ & Fluency $\uparrow$ \\
& Agreement / $\kappa$ & Agreement / $\kappa$  \\
\midrule
DEVTC & 63.5 / 0.69 & 86.5 / 0.80 \\
R2D2 & 71.5 / 0.73 & \textbf{90.0} / 0.84 \\
\ours & \textbf{75.0} / 0.76 & 88.0 / 0.81 \\
\bottomrule
\end{tabular}
\caption{Human evaluation results on the criteria of faithfulness and fluency, with the total agreement by Fleiss’ Kappa ($\kappa$)~\cite{fleiss1971measuring}. \ours has the best performance in terms of faithfulness, while achieving comparable performance in fluency.}
\label{tab:human_faith}
\end{table}

\section{Experimental Setup}
We next discuss the evaluation metrics, baselines, and implementation details for the experiments.
\subsection{Evaluation Metrics}
We applied various automated evaluation metrics at different levels to evaluate the model performance from multiple perspectives.
\paragraph{Surface-level}
Following \citet{chen-etal-2020-logical}, we used BLEU-1/2/3 to measure the consistency of generated statements with the reference. 

\paragraph{Diversity-level}
We used Distinct-$n$~\cite{li-etal-2016-diversity} and self-BLEU-$n$~\cite{self-bleu} to measure the diversity of five generated statements for each table. Distinct-$n$ is defined as the total number of distinct $n$-grams divided by the total number of tokens in the five generated statements; Self-BLEU-$n$ measures the average $n$-gram BLEU score between generated statements. We measured Distinct-$2$ and Self-BLEU-$4$ in our experiment.

\paragraph{Faithfulness-level}
Similar as the previous works~\cite{chen-etal-2020-logical,nan2022r2d2, liu2022plog}, we used a parsing-based evaluation metric (i.e., SP-Acc) and two NLI-based evaluation metrics (i.e., NLI-Acc and TAPEX-Acc) to measure the faithfulness of generation. 
SP-Acc directly extracts the meaning representation from the generated sentence and executes it against the table to verify the correctness.
NLI-Acc and TAPEX-Acc use TableBERT~\cite{Chen2020TabFact} and TAPEX~\cite{liu2022tapex} respectively as their backbones, and were finetuned on the \textsc{TabFact} dataset~\cite{Chen2020TabFact}.
\citet{liu2022plog} found that NLI-Acc is overly positive about the predictions, while TAPEX-Acc is more reliable to evaluate the faithfulness  of generated sentences.

\subsection{Baseline Systems}
We implemented following baseline systems for the performance comparison:
\textbf{GPT2-TabGen}~\cite{chen-etal-2020-logical} directly fine-tunes GPT-2 over the \logicnlg dataset;
\textbf{GPT2-C2F}~\cite{chen-etal-2020-logical} first produces a template which determines the global logical structure, and then generates the statement conditioned on the template;
\textbf{DCVED}~\cite{chen-etal-2021-de} applies a de-confounded variational encoder-decoder to reduce the spurious correlations during LT2T generation training;
\textbf{DEVTC}~\cite{perlitz2022diversity} utilized reasoning operation types as an explicit control to increase the diversity of LT2T generation;
and \textbf{R2D2}~\cite{nan2022r2d2} trains a generative language model both as a generator and a faithfulness discriminator with additional replacement detection and unlikelihood learning tasks, to enhance the faithfulness of LT2T generation.

\subsection{Implementation Details} \label{sec:setting}
Following \citet{shu-etal-2021-logic}, we converted each logic form into a more human-readable form for both \ours training and inference data.
\ours was implemented using fairseq library~\cite{fairseq}, with BART-Large~\cite{lewis-etal-2020-bart} as the backbones. All experiments were conducted on an 8 NVIDIA RTX-A5000 24GB cluster. Both \ours and the statement verifier was trained for 5,000 steps with a batch size of 128. The best checkpoints were selected by the validation loss. 

\section{Experimental Results}
This section discusses automated and human evaluation results of different systems.
\subsection{Main Results}
Table~\ref{tab:logicnlg-auto} presents the results on \logicnlg.
\ours outperforms all the baselines on the criteria of diversity and faithfulness, and is the first model that achieves state-of-the-art results on both faithfulness- and diversity-level.
It is worth noting that in the \logicnlg setting, a generated statement is allowed to cover a different table region or reasoning operations from the references, as long as it is fluent and factually correct. However, in such cases, the reference-based metrics will be low, explaining why the BLEU-1/2/3 scores of \ours are lower than other models. 

\subsection{Human Evaluation}
We conducted the human evaluation with four expert annotators using the following three criteria: (1) \emph{Faithfulness} (scoring 0 or 1): if all facts contained in the generated statement are entailed by the table content; (2) \emph{Diversity} (voting the best \& worst): if the five generated statements cover information from different table regions, and use different reasoning operations; (3) \emph{Fluency} (scoring 0 or 1): if the five generated statements are fluent and without any grammar mistakes.

We chose R2D2~\cite{nan2022r2d2} and DEVTC~\cite{perlitz2022diversity} for comparison, as they achieved best-performance results in faithfulness and diversity, respectively.  We sampled 50 tables from the \logicnlg test set. For each table, we selected all five generated statements from each model's output. To ensure fairness, the model names were hidden to the annotators, and the display order between three models was randomly shuffled. Human evaluation results show that \ours delivers improvements in both faithfulness (Table~\ref{tab:human_faith}) and diversity (Table~\ref{tab:human-study-vanilla}), while achieving comparable performance in fluency (Table~\ref{tab:human_faith}).

\section{Conclusions}
This work proposes \ours, which utilizes logic forms as fact verifiers and content planners to enable controllable LT2T generation.  Experimental results on \logicnlg demonstrate that \ours delivers a great improvement in both diversity and faithfulness of LT2T generation.
\section*{Limitations}
The first limitation of our approach is that \ours does not explore long text generation~\cite{sciGen}. \ours only supports the generation of multiple single sentences. To enable long text generation (i.e., generate a long paragraph that delivers various perspectives on the table data), a global content planner~\cite{su-etal-2021-plan-generate} needs to be designed to highlight which candidate sentences should be mentioned and in which order. Additionally, we believe that \ours can also be applied to text generation over hybrid context with both textual and tabular data~\cite{hybridqa,zhao2022multihiertt,nakamura-etal-2022-hybridialogue}. 

The second limitation of our work is that the statement verifier discussed in Section~\ref{sec:verifier} was trained using the same data as NLI-Acc and TAPEX-Acc. This might bring some bias for NLI-based metrics on faithulness-level evaluation. In the future, we will exploit a more robust automated evaluation system~\cite{fabbri-etal-2021-summeval, rose} to comprehensively evaluate the LT2T model performances from different perspectives.

Moreover, we applied the SASP model~\cite{ou-liu-2022-learning} to convert statements into logic forms (Section~\ref{sec: train_data}). Some converted logic forms may be inconsistent with the original statement. We believe that future work could incorporate the Logic2Text~\cite{chen_2020_logic2text} dataset into training data to further improve the \ours performance.

\section*{Ethical Consideration}
We used the \logicnlg~\cite{chen-etal-2020-logical} dataset for training and inference. \logicnlg is publicly available under MIT license\footnote{\url{https://opensource.org/licenses/MIT}} and widely used in NLP research and industry. 

\bibliography{anthology,custom}
\bibliographystyle{acl_natbib}
\appendix
\section{Appendix}
\begin{table*}[htb]
\centering
\setlength{\extrarowheight}{3pt}
\resizebox{0.98\linewidth}{!}{
\begin{tabular}{l|l|l|l|l|l}
\toprule
Reasoning Op                   & Function Category                           & Name                          & Arguments                   & Output & Description                                                                                        \\ \midrule
Unique                       &    UNIQUE                                & only                          & view                        & bool   & returns whether there is exactly one row in the view                                               \\  \midrule
Aggregation                  & AGGREGATION                        & avg/sum                       & view, header, string        & number & returns the average/sum of the values under the header column                                      \\ \midrule
Count & COUNT & count                         & view                        & number & returns the number of rows in the view                                                             \\ \midrule

\multirow{3}{*}{Ordinal}    & ORD\_ARG                           & nth\_argmax/nth\_argmin       & view, header string         & view    & returns the row with the n-th max/min value in header column                                       \\ \cline{2-6} 
& ORDINAL                            & nth\_max/nth\_min             & view, header string         & number & returns the n-th max/n-th min of the values under the header column                                \\ \cline{2-6} 
& SUPER\_ARG                         & argmax/argmin                 & view, header string         & view    & returns the row with the max/min value in header column                                            \\ \midrule

\multirow{4}{*}{Comparative} & \multirow{4}{*}{COMPARE}           & eq/not\_eq                    & object, object              & bool   & returns if the two arguments are equal                                                             \\ \cline{3-6} 
                             &                                    & round\_eq                     & object, object              & bool   & returns if the two arguments are roughly equal under certain tolerance                             \\ \cline{3-6} 
                             &                                    & greater/less                  & object, object              & bool   & returns if 1st argument is greater/less than 2nd argument                                              \\ \cline{3-6} 
                             &                 & diff                          & object, object              & object & returns the difference between two arguments                                                       \\ \midrule
\multirow{6}{*}{Majority}    & \multirow{6}{*}{MAJORITY}          & all\_eq/not\_eq               & view, header string, object & bool   & returns whether all the values under the header column are equal/not equal to 3rd argument           \\ \cline{3-6} 
                             &                                    & all\_greater/less             & view, header string, object & bool   & returns whether all the values under the header column are greater/less than 3rd argument            \\ \cline{3-6} 
                             &                                    & all\_greater\_eq/less\_eq     & view, header string, object & bool   & returns whether all the values under the header column are greater/less or equal to 3rd argument     \\ \cline{3-6} 
                             &                                    & most\_eq/not\_eq              & view, header string, object & bool   & returns whether most of the values under the header column are equal/not equal to 3rd argument       \\ \cline{3-6} 
                             &                                    & most\_greater/less            & view, header string, object & bool   & returns whether most of the values under the header column are greater/less than 3rd argument        \\ \cline{3-6} 
                             &                                    & most\_greater\_eq/less\_eq    & view, header string, object & bool   & returns whether most of the values under the header column are greater/less or equal to 3rd argument \\ \midrule
                             
\multirow{4}{*}{Conjunction}      & \multirow{3}{*}{FILTER}            & filter\_eq/not\_eq            & view, header string, object & view   & returns the subview whose values under the header column is equal/not equal to 3rd argument          \\ \cline{3-6} 
                             &                                    & filter\_greater/less          & view, header string, object & view   & returns the subview whose values under the header column is greater/less than 3rd argument           \\ \cline{3-6} 
                             &                                    & filter\_greater\_eq /less\_eq & view, header string, object & view   & returns the subview whose values under the header column is greater/less or equal than 3rd argument  \\ \cline{2-6} 
                             & OTHER              & filter\_all                   & view, header string         & view   & returns the view itself for the case of describing the whole table                                 \\\midrule

                    \multirow{2}{*}{Other}         &   OTHER          & hop                           & view, header string          & object & returns the value under the header column of the row                                               \\ \cline{2-6} 
                             &        OTHER                            & and                           & bool, bool                  & bool   & returns the boolean operation result of two arguments                                              \\
\bottomrule
\end{tabular}
}

\caption{A complete list of function definitions for the logic forms (Similar as \citet{chen_2020_logic2text}).}

\label{table:app_func}
\end{table*}

\end{document}